\documentclass[a4paper,twoside]{article}

\usepackage{microtype}
\usepackage[hidelinks]{hyperref}
\usepackage{url}
\usepackage{booktabs}

\usepackage{graphicx}

\usepackage[outdir=./]{epstopdf}

\usepackage{epsfig}
\usepackage{subcaption}
\usepackage{calc}
\usepackage{amssymb}
\usepackage{amstext}
\usepackage{amsmath}
\usepackage{amsthm}
\usepackage{multicol}
\usepackage{pslatex}
\usepackage{apalike}
\usepackage{algorithm2e}
\usepackage[bottom]{footmisc}

\usepackage{SCITEPRESS}     

\begin{document}

\title{Enhancing Answer Attribution for Faithful Text Generation with Large Language Models}

\author{\authorname{Juraj Vladika\orcidAuthor{0000-0002-4941-9166}, Luca Mülln and Florian Matthes\orcidAuthor{0000-0002-6667-5452}}
\affiliation{Technical University of Munich, School of Computation, Information and Technology, Department of Computer Science, Germany}
\email{ \{juraj.vladika, luca.muelln, matthes\}@tum.de }
}


\abstract{The increasing popularity of Large Language Models (LLMs) in recent years has changed the way users interact with and pose questions to AI-based conversational systems. An essential aspect for increasing the trustworthiness of generated LLM answers is the ability to trace the individual claims from responses back to relevant sources that support them, the process known as \textit{answer attribution}. While recent work has started exploring the task of answer attribution in LLMs, some challenges still remain. In this work, we first perform a case study analyzing the effectiveness of existing answer attribution methods, with a focus on subtasks of answer segmentation and evidence retrieval. Based on the observed shortcomings, we propose new methods for producing more independent and contextualized claims for better retrieval and attribution. The new methods are evaluated and shown to improve the performance of answer attribution components. We end with a discussion and outline of future directions for the task.
}

\onecolumn \maketitle \normalsize \setcounter{footnote}{0} \vfill

\section{\uppercase{Introduction}}
As Large Language Models (LLMs) rise in popularity and increase their capabilities for various applications, the way users access and search for information is noticeably changing \cite{kaddour2023challenges}.
The impressive ability of LLMs to produce human-sounding text has led to new applications but also raised concerns. They sometimes generate responses that sound convincing but lack accuracy or credible sources, so-called hallucinations \cite{ji2023survey}. This poses challenges to their reliability, especially in critical applications like law or healthcare, as well as in day-to-day usage \cite{wang2024decodingtrust}. 

Additionally, the opaque nature of these models complicates understanding their decision-making processes and interpretability of generated outputs \cite{singh2024rethinking}.
As these models continue to permeate various sectors, from education \cite{kasneci2023chatgpt} to healthcare \cite{nori2023capabilities} — the need for verifiable and accountable information becomes increasingly crucial. 
If LLMs provide incorrect information or biased content, the inability to trace back the origin of such responses can lead to misinformation and potential harm or infringe on copyrighted material \cite{MorganLewis2023GenAI}. 

A promising avenue for increasing the trustworthiness and transparency of LLM responses is the idea of \textit{answer attribution}. It refers to the process of tracing back ("attributing") the claims from the output to external evidence sources and showing them to users \cite{rashkin2023measuring}. 
Distinct from usual methods of hallucination mitigation, which focus on altering the model's output, answer attribution is oriented towards end users. It aims to equip users with a list of potential sources that support the output of the LLM to increase its transparency and leaves quality assurance to the users. This process usually involves segmenting LLM answers into claims and linking them to relevant evidence. While many attribution systems have started emerging in recent years \cite{li2023survey}, we observe they still suffer from drawbacks limiting their applicability. The retrieved sources for specific claims and their respective entailment can be inaccurate due to inadequate claim formulation \cite{liu-etal-2023-evaluating,min-etal-2023-factscore}.



To address these research gaps, in this study, we provide incremental contributions to the answer attribution process by enhancing its components. We: (1) perform a case study of current answer attribution components from literature and detect their shortcomings; (2) propose improvements to the answer-segmentation and evidence-retrieval components; and (3) provide a numerical and qualitative analysis of improvements. We involve human annotation on subsets when possible and consider multiple competing approaches. Our research builds on top of recent LLM factuality and answer attribution works and outlines open challenges, leaving the door open for further advancements and refinement of the process.




\section{\uppercase{Related Work}}
A lot of ongoing NLP work is devoted to ensuring the trustworthiness of LLMs in their everyday use \cite{liu2024trustworthy}, including their reliability \cite{zhang2023chatgpt}, safety \cite{wei2024jailbroken}, fairness \cite{li2023survey}, efficiency \cite{afzal23}, or explainability \cite{zhao2024explainability}. An important aspect hindering the trust in LLMs are hallucinations -- described as model outputs that are not factual, faithful to the provided source content, or overall nonsensical \cite{ji2023survey}. 

A recent survey by \cite{zhang2023sirens} divides hallucinations into input-conflicting, context-conflicting, and fact-conflicting. Our work focuses on fact-conflicting, which are hallucinations in which facts in output contradict the world knowledge. 
Detecting hallucinations is tied to the general problem of measuring the factuality of model output \cite{augenstein2023factuality,zhao2024felm} and automated fact-checking of uncertain claims \cite{guo2022survey,vladika-matthes-2023-scientific}. The recently popular method FactScore evaluates factuality by assessing how many atomic claims from a model output are supported by an external knowledge source \cite{min-etal-2023-factscore}. Hallucinations can be corrected in the LLM output by automatically rewriting those claims found to be contradicting a trusted source, as seen in recent CoVe \cite{dhuliawala2023chainofverification} or Factcheck-Bench \cite{wang2024factcheckbenchfinegrainedevaluationbenchmark}. 

A middle ground between pure factuality evaluation and fact correction is answer attribution. The primary purpose of answer attribution is to enable users to validate the claims made by the model, promoting the generation of text that closely aligns with the
cited sources to enhance accuracy \cite{li2023survey}. One task setting is evaluating whether the LLMs can cite the references for answers from their own memory \cite{bohnet2023attributed}. A more common setup involves retrieving the references either before the answer generation or after generating it \cite{malaviya2024expertqa}. When attributing claims to scientific sources, the more recent and better-cited publications were found to be the most trustworthy evidence \cite{vladika-matthes-2024-improving}. Some approaches to the problem include fine-tuning smaller LMs on NLP datasets \cite{yue-etal-2023-automatic} or using human-in-the-loop methods \cite{hagrid}. Our work builds on top of \cite{malaviya2024expertqa} by utilizing their dataset but improves the individual components of the attribution pipeline.



\section{\uppercase{Foundations}}
We provide a precise description for the task of attribution in the context of LLMs for this work as follows: \textbf{Answer Attribution} is the task of providing a set of sources $s$ that inform the output response $r$ of a language model for a given query $q$. These sources must be relevant to the model's response and should contain information that substantiates the respective sections of the response. 
This definition provides a comprehensive overview of the task and encapsulates its constituent subtasks:
\begin{enumerate}
\small
    \item \textbf{Response Segmentation}: Segmenting the response $r$ into individual claims $c_i$.
    \item \textbf{Claim Relevance Determination}: Determining the relevance of each claim $c_i$ for the need of attribution ("claim check-worthiness").
    \item \textbf{Finding Relevant Evidence}: Retrieving a list of relevant evidence sources $s_i$ for each claim $c_i$.
    \item \textbf{Evidence-Claim Relation}: Determining whether the evidence sources from the list of sources $s_i$ actually refer to the claim $c_i$.\\
\end{enumerate}


In our work, we focus on analyzing and improving subtasks 1 and 4, and to a lesser extent, subtasks 2 and 3, leaving further improvements to future work. We take the recent dataset \textit{ExpertQA} \cite{malaviya2024expertqa} as a starting point for our study. Moving away from short factoid questions, this dataset emulates how domain experts in various fields interact with LLMs. Thus, the questions posed to the model are longer and more complex, can contain hypothetical scenarios, and elicit long, descriptive responses. This makes it a realistic benchmark for modern human-LLM interaction.

\begin{figure}[htpb]
        \includegraphics[width=0.45\textwidth]{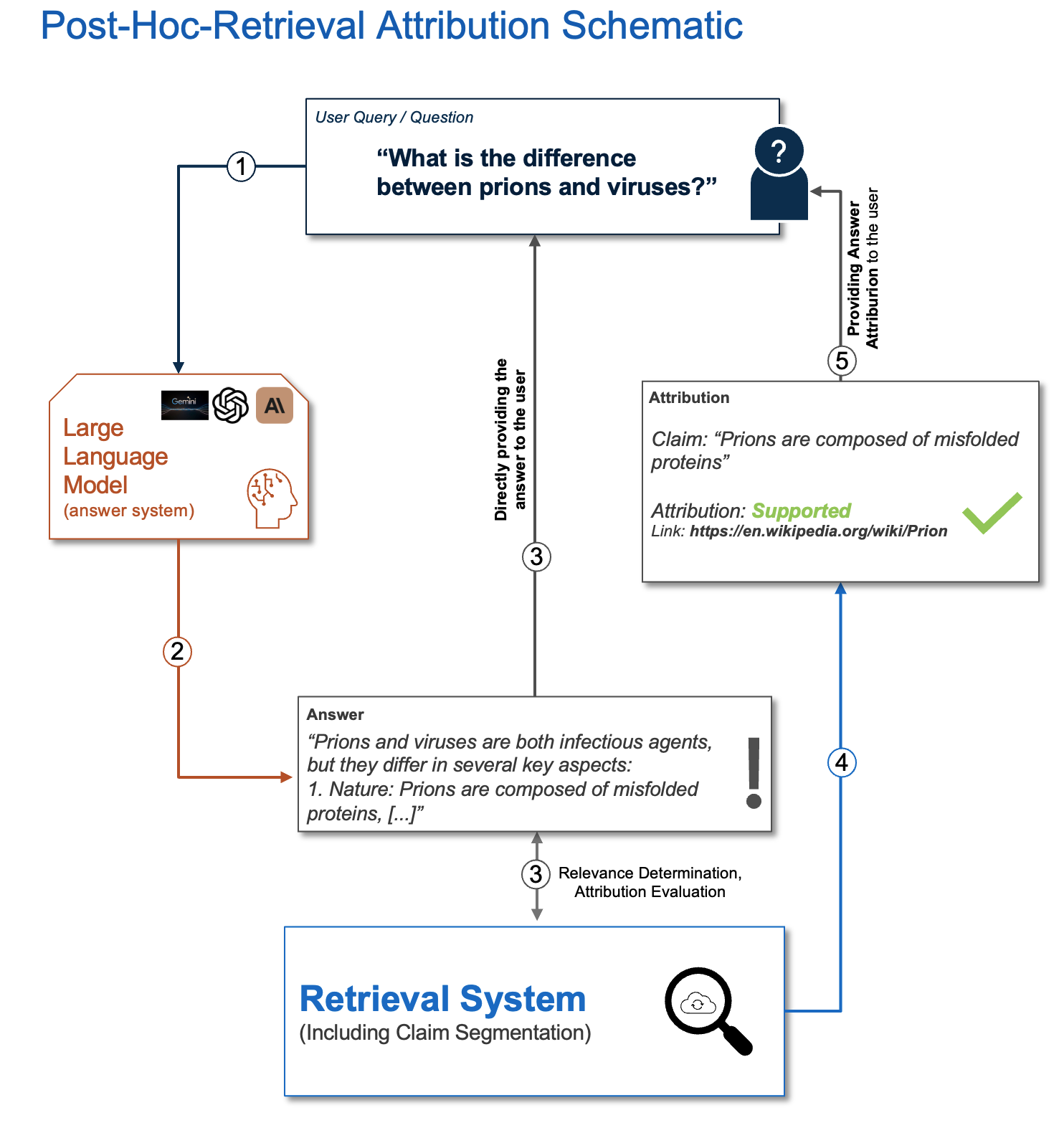} 
        \caption{The complete answer attribution process (in the Post-Hoc-Retrieval setup)}
        \label{fig:phr}
    \vfill
\end{figure}

We take the responses generated by GPT-4 ("gpt-4" in OpenAI API) from ExpertQA and perform attribution evaluation based on claims found in its responses. Two main setups for attribution are post-hoc retrieval (PHR), which first generates the response and then does retrieval to attribute the facts; and retrieve-then-read (RTR), which first retrieves the sources and then generates the response (i.e., RAG). 
In our work, we focus on the PHR system (Fig. \ref{fig:phr}), because it is closer to the definition of attribution. Still, the challenges in claim formulation and evidence retrieval apply to both settings, so our findings also hold for RTR. 

\begin{table*}[htpb]
\small
    \centering
        \caption{High-level comparison of the different answer segmentation systems.}
    \begin{tabular}{lcccc}
    \hline
    \textbf{Segmentation System} & \textbf{Number of $c$} & \textbf{Unique \#$c$} & \textbf{avg. len($c$)} & \textbf{$c$ / Sentence}  \\ \hline
    spaCy\_sentences & 938 & 855 & 103.2 & 1.00 \\
    gpt35\_factscore & 3016 & 2684 & 61.4 & 3.2 \\
    segment5\_propsegment & 2676 & 2232 & 54.2 & 2.85 \\
     \hline
    \end{tabular}
    \label{tab:claim_segmentation_high_level}
\end{table*}


\section{\uppercase{Case Study of Existing Solutions}}
This section provides a case study of recently popular approaches for different components of the answer attribution pipeline.

\subsection{Answer Segmentation}

As described above, the first step for attribution in PHR systems is to segment the provided LLM response into claims (atomic facts). We define a claim as "a statement or a group of statements that can be attributed to a source". The claim is either a word-by-word segment of the generated answer or semantically entailed by the answer. 
To validate the segmentation, we sample 20 random questions from the ExpertQA dataset. Three different segmentation systems are evaluated based on the number of atomic facts each claim contains and the number of claims they generate. 

The first (\textbf{i}) and most intuitive way of segmenting an answer into claims is to use the syntactic structure of the answer, segmenting it into sentences, paragraphs, or other syntactic units. Following ExpertQA \cite{malaviya2024expertqa}, this segmentation is done using the sentence tokenizer from the Python library \texttt{spaCy}.\footnote{\url{https://spacy.io/}} The second approach (\textbf{ii}) for answer segmentation that we analyze is based on the work of PropSegment \cite{chen2023propsegment}, where text is segmented into \textit{propositions}. A proposition is defined as a unique subset of tokens from the original sentence. We use the best-performing model from the paper, SegmenT5-large \cite{chen2023subsentence}, a fine-tuned version of the T5 checkpoint 1.1 \cite{chung2022scaling}. 
The third approach (\textbf{iii}) of segmenting an answer into claims utilizes pre-trained LLMs and prompting, as found in FactScore \cite{min-etal-2023-factscore}. In their approach, the model is prompted to segment the answer into claims, and the resulting output is subsequently revised by human annotators. 
We replicate this method by using GPT-3.5 (turbo-0613) and the same prompt (\textit{"Please breakdown the following sentence into independent facts:"}), amended with meta-information and instructions for the model on formatting the output. The prompt is in Appendix \ref{sec:appendix-prompts}, Table \ref{tab:checkworth}. 

Table \ref{tab:claim_segmentation_high_level} shows the differences between the three answer segmentation approaches. As expected, the average number of characters of the atomic facts created by GPT-3.5 and T5 is significantly smaller than the original sentence length. It is also noteworthy that the claims generated by GPT-3.5 are longer in characters and more numerous per sentence. In addition, the number of unique claims per answer and the number of claims per answer differ significantly by an average of 12\% and up to 16.5\% for SegmenT5. An error we observed is that the segmentation systems create duplicated claims for the same answer.

For a qualitative analysis of these segmented claims, we manually annotate 122 claims that the three systems generated for a randomly selected question \textit{"A 55 year old male patient describes the sudden appearance of a slight tremor and having noticed his handwriting getting smaller, what are the possible ways you'd find a diagnosis?"}. The categories for annotations are aligned with \cite{chen2023propsegment} and \cite{malaviya2024expertqa}, and describe important claim properties. The properties are as follows: (1) \textbf{Atomic}: the claim contains a single atomic fact; (2) \textbf{Independent}: the claim can be verified without additional context; (3) \textbf{Useful}: the claim is useful for the question; (4) \textbf{Errorless}: the claim does not contain structural errors, e.g., being an empty string; (5) \textbf{Repetition}: the claim is a repetition of another claim from the same segmentation system. Each category is binary, meaning a claim can be annotated with multiple categories. 
Given that the question is from the medical domain, the claims are expected to be more complex and require domain knowledge.

\begin{table*}[htpb]
\small
    \centering
     \caption{Comparison of different claim properties for the different segmentation systems. The fractions show the number of occurrences divided by the total number of atomic claims generated by that system.}
    \begin{tabular}{@{}lccccc@{}}
    \toprule
    & \textbf{Atomic} & \textbf{Independent} & \textbf{Useful} & \textbf{Errorless} & \textbf{Repetition} \\ \midrule
    gpt35\_factscore & 53/56  & 8/56 & 44/56  & 48/56 & 13/56      \\
    segment5\_propsegment & 40/53  & 6/53 & 28/53  & 34/53 & 18/53      \\
    spaCy\_sentences & 3/15 & 3/15 & 10/15  & 11/15 & 3/15       \\ \bottomrule
    \end{tabular}
       
    \label{tab:comparison_scores}
\end{table*}

Table \ref{tab:comparison_scores} shows the result of the qualitative analysis. The most noticeable outcome is that the \texttt{spaCy} segmentation system performs significantly differently compared to other systems. It simply tokenizes the responses into sentences and considers every sentence to be a claim, which is not realistic given the often quite long sentences generated by LLMs. Consequently, the score for "Atomic" claims stands at 20\% (3/15). Intriguingly, only 20\% (3/15) of the sentences from the response are independently verifiable without additional context from the question or the rest of the response. Due to the complexity of the answer, most sentences reference a preceding sentence in the response, mentioning "the patient" or "the symptoms".

The usefulness of the claims in answering the given questions is relatively high for spaCy sentence segmentation and GPT-3.5 segmentation but diminishes for the SegmenT5 segmentation. Although most claims are errorless, it is notable that all systems produce erroneous outputs. Specifically, for this question, \texttt{spaCy} segments four empty strings as individual sentences. It is plausible that errors in the other two segmentation systems stem from this issue, as they also rely on \texttt{spaCy}-tokenized sentences as input. This dependency also results in repetitions, primarily based on incorrect answer segmentation. 
This list provides a positive and a negative example claim for each category to give an idea of errors:

\begin{enumerate}\label{enum:claim_examples}
\small
    \item \textbf{Atomic} |  
    \textit{Positive:} "Seeking a second opinion helps" (\texttt{gpt35\_factscore}) --
    \textit{Negative:} "Brain tumors or structural abnormalities are among the possible causes that these tests aim to rule out." (\texttt{gpt35\_factscore})
    \item \textbf{Independent} |
    \textit{Positive}: "Parkinson's disease is a cause of changes in handwriting." (\texttt{segment5\_propsegment}) --
    \textit{Negative}: "Imaging tests may be ordered." (\texttt{segment5\_propsegment})
    \item \textbf{Useful} |
    \textit{Positive}: "There are several possible diagnoses that could explain the sudden appearance of a slight tremor and smaller handwriting." (\texttt{gpt35\_factscore}) --
    \textit{Negative}: "The patient is a 55-year-old male." (\texttt{segment5\_propsegment})
    \item \textbf{Errorless} | 
    \textit{Positive}: "The patient is experiencing smaller handwriting." (\texttt{gpt35\_factscore}) --
    \textit{Negative}: "The sentence is about something." (\texttt{segment5\_propsegment})
\end{enumerate}

Based on these findings, we conclude that automatic answer segmentation faces three main challenges and we give three desiderata for successful answer segmentation: (1) To provide independently verifiable claims, the segmentation system requires more context than just the sentence, possibly the entire paragraph and the question; (2) the segmentation system needs to be capable of handling domain-specific language, such as the complex medical domain; (3) if the goal is to identify individual atomic facts, the segmentation system needs to operate at a more granular level than sentences.

\subsection{Claim Relevance}
\label{sec:claim_relevance_determination}
The relevance (usefulness) of a claim is evaluated based on its relation to the question. We define it as: \textit{Given a question or query $q$ and a corresponding answer $a$, a claim $c$ with $c \in a$ is relevant if it provides information to satisfy the user’s information need}. Most attribution publications do not perform the relevance evaluation automatically, relying instead on annotators \cite{min-etal-2023-factscore}. Due to limited resources, we want to investigate whether this can be performed automatically. We adopt the approach of FactCheck-Bench \cite{wang2024factcheckbenchfinegrainedevaluationbenchmark}, who implement it with a GPT-3.5 prompt -- the prompt is in Appendix \ref{sec:appendix-prompts}, Table \ref{tab:checkworth}. They classify a claim into four classes of "check-worthiness": factual claim, opinion, not a claim (e.g., \textit{Is there something else you would like to know?}), and others (e.g., \textit{As a language model, I cannot access personal information}). 

To evaluate the performance, we use the same 122 claims from Table \ref{tab:comparison_scores} and annotate with the LLM and manually. 
The agreement for "factual claim" class is very high (79 annotations the same out of 85), while the biggest confusion is between "not a claim" and "other". This shows that automatic assessment can reliably be used to determine the claim relevance. Therefore, we apply the prompt to automatically label all the claims from Table \ref{tab:claim_segmentation_high_level}. The results are shown in Table \ref{tab:claim_segmentation_low_level}. We observe that 86.3\% claims generated by GPT3.5 FactScore system are factual. These 2,317 claims will be used in further steps for attribution evaluation.

\begin{table*}[htpb]
\small
    \centering
    \caption{Claim relevance distribution of different segmentation systems.}
    \begin{tabular}{lccccc}
    \hline
    \textbf{Segmentation System} & \textbf{Unique \#$c$} & \textbf{\# factual} & \textbf{\# not a claim} & \textbf{\# opinion} & \textbf{\# other}\\ 
    \hline
    spaCy\_sentences & 855 & 550 & 244 & 26 & 35 \\
    gpt35\_factscore & 2684 & 2317 & 258 & 68 & 41 \\
    segment5\_propsegment & 2232 & 1878 & 290 & 36 & 28 \\
    \hline
    \end{tabular}
    
    \label{tab:claim_segmentation_low_level}
\end{table*}

\subsection{Evidence Retrieval}
\label{sec:information_retrieval_impl}
The evidence retrieval step in the attribution process is arguably the most important, especially in a post-hoc retrieval system -- its goal is to find the evidence to which a claim can be attributed to.
Evidence sources can be generated directly from LLM's memory \cite{ziems-etal-2023-large}, retrieved from a static trusted corpus like Sphere \cite{piktus2021web} or Wikipedia \cite{peng2023check}, or dynamically queried from Google 
\cite{gao-etal-2023-rarr}.
We use the Google approach: we take each claim (labeled as unique and factual in the previous steps) and query Google with it, take the top 3 results, scrape their entire textual content from HTML, and split it (with \textit{NLTK}) into chunks of either 256 or 512 character length. We embed each chunk with a Sentence-BERT embedder \textit{all-mpnet-base-v2} \cite{mpnet} and store the chunks into a FAISS-vector database \cite{douze2024faiss}. After that, we query each claim against the vector store for that question and retrieve the top 5 most similar chunks.

\begin{table}[htpb]
    \centering
    \caption{NLI predictions between a claim and its respective evidence snippets found on Google.}
    \begin{tabular}{lccc}
    \hline
    \textbf{Method \& CW} & \textbf{Contr.} & \textbf{Entail.} & \textbf{No Relation}  \\ \hline
    GPT3.5 - 256& 2 & 111 & 82 (36.0\%)\\
    GPT3.5 - 512& 1& 126 & 88 (38.6\%) \\ 
    DeBERTa - 256& 12 & 37 & 179 (78.5\%)\\ 
    DeBERTa - 512& 11 & 64& 153 (67.1\%) \\ 
    Human - 256& 8 & 54 & 166 (72.8\%) \\ 
    Human - 512& 9 & 81 & 138 (60.5\%) \\ 
    \hline
    \end{tabular}
    
    \label{tab:claim_based_Google_search}
\end{table}

We want to automatically determine whether the retrieved evidence chunk is related to the claim. We model this as a Natural Language Inference (NLI) task, following the idea from SimCSE \cite{gao2021simcse}, where two pieces of text are semantically related if there is an entailment or contradiction relation between them and unrelated otherwise. For this purpose, we use GPT-3.5 with a few-shot prompt (Appendix \ref{sec:appendix-prompts}, Table \ref{tab:gpt_nli}) and DeBERTa-v3-large model fine-tuned on multiple NLI datasets from \cite{Laurer_van_Atteveldt_Casas_Welbers_2024}, since DeBERTa was shown to be the most powerful encoder-only model for NLI. 

We take 228 claim-evidence pairs and annotate them both manually and automatically with the two models (GPT and DeBERTa). The results are in Table \ref{tab:claim_based_Google_search}. The results show that the DeBERTa-NLI model was by far more correlated with human judgment and that GPT-3.5 was overconfident in predicting the entailment relation, i.e., classifying a lot of irrelevant chunks as relevant. Additionally, the longer context window led to these longer evidence chunks being more related to the claim. The stricter nature of DeBERTa predictions makes it better suited for claim-evidence relation prediction. Therefore, we will use DeBERTa as the main NLI model in the next section, with a 512-character context window. 


\section{\uppercase{Developing Solutions}}\label{sec:developing_solutions}
In this section, we propose certain solutions for selected key issues identified in the previous section. We use the existing answer attribution pipeline and enhance individual components to assess their effects on the overall system.

\subsection{Answer Segmentation}\label{sec:claim_segmentation_impl}
One of the primary reasons for the weak performance of previous systems was the lack of independence among claims. Even when tasked to create atomic claims, most existing systems fail to provide sufficient context, making it difficult for the claims to stand alone. This leads to significant error propagation and misleading outcomes in evidence retrieval and attribution evaluation. 
There are three different types of claims produced by current systems that require additional context for accurate evaluation:
\begin{enumerate}
\scriptsize
    \item \textbf{Anaphoric References (Coreference Resolution)}: Claims that include one or more anaphors referring to previously mentioned entities or concepts. |  
    \textbf{Example}: "The purpose of \textit{these strategies} is to reduce energy consumption.", "\textit{They} ensure the well-being of everyone."
    \item \textbf{Conditioning (Detailed Contextualization)}: Claims that lack entire sentences or conditions necessary for proper contextualization. While not always obvious from the claim itself, this information is crucial for accurately evaluating the claims. | 
    \textbf{Example:} "Chemotherapy is no longer the recommended course of action."
    \item \textbf{Answer Extracts (Hypothetical Setup)}: Claims that arise from questions describing a hypothetical scenario. Current answer systems often replicate parts of the scenario in the answer, leading to claims that cannot be evaluated independently of the scenario itself. | 
    \textbf{Example:} "A young girl is running in front of cars."
\end{enumerate}

We propose two strategies to provide more context during answer segmentation: (1) claim enrichment, and (2) direct segmentation with context. In the first approach, we edit extracted claims to incorporate the necessary context from both the answer and the question. A system employing this strategy would implement the function $f_{\text{enrich}}(q, r, c_{\text{non-independent}})$, where $c_{\text{non-independent}}$ is the non-independent claim, $r$ is the response, and $q$ is the question. In the second approach, we suggest a system that directly segments the answer into multiple independent claims, each supplemented with the required context. This system would use the function $f_{\text{segment}}(r, q)$, differing from the initial systems (as in Section 4), by incorporating the entire answer and question rather than basing the segmentation on individual sentences. 

\subsubsection{Claim Enrichment}
We want to enrich only the non-independent claims. In the previous section, we manually labeled the claims for independence (Table \ref{tab:comparison_scores}). We now want to automate this task. For this purpose, we test whether the GPT-3.5 (turbo-0613) and GPT-4 (turbo-1106) systems can perform this task with a one-shot prompt (in Appendix \ref{sec:appendix-prompts}, Table \ref{tab:enrich_prompt}) that assesses the independence. The results are compared with human evaluation from Table \ref{tab:comparison_scores}. Table \ref{tab:non_independence_detection} shows the results. It is evident that both GPT-3.5 and GPT-4 exhibit significantly high precision, with GPT-4 outperforming in terms of recall and F1 score. We conclude that claim independence can be detected by LLMs (0.84 F1 in GPT4) and utilize the claims classified as "non-independent" by GPT-4 to assess the performance of the function $f_{\text{enrich}}(q, r, c_{\text{non-independent}})$.

\begin{table}[htpb]
\scriptsize
    \centering
    \caption{Non-Independence detection performance compared to human evaluation. 
    }
    \begin{tabular}{l|ccc|ccc}
        \hline
        & \multicolumn{3}{c}{GPT3.5} & \multicolumn{3}{c}{GPT4} \\
        \cline{2-7}
        System & Prec. & R & F1 & Prec. & Rec. & F1 \\
        \hline
        \textbf{Overall} & 0.94 & 0.27 & 0.42 & \textbf{0.96} & \textbf{0.74} & \textbf{0.84} \\
        factscore & 0.93 & 0.29 & 0.44 & 0.95 & 0.75 & 0.84 \\
        segmenT5 & 0.90 & 0.19 & 0.32 & 0.96 & 0.66 & 0.78 \\
        spaCy & 1.0 & 0.5 & 0.67 & 1.0 & 1.0 & 1.0 \\
        \hline
    \end{tabular}
    \label{tab:non_independence_detection}
\end{table}

\begin{table*}[htpb]
\small
    \centering
    \caption{Descriptive comparison of adopted answer segmentation approaches.}
    \begin{tabular}{lcccc}
    \hline
    \textbf{Segmentation System} & \textbf{Number of $c$} & \textbf{Unique \#$c$} & \textbf{avg. len($c$)} & \textbf{$c$ / Sentence}  \\ \hline
    \textbf{GPT3.5 direct} & 644 & 644 & 102.8 & 0.75 \\
    \textbf{GPT4 direct} & 948 & 948 & 84.1 & 1.11 \\
   \textit{ spaCy\_sentences} & 938 & 855 & 103.2 & 1.00 \\
  \textit{  gpt35\_factscore} & 3016 & 2684 & 61.4 & 3.22 \\
     \hline
    \end{tabular}
    
    \label{tab:claim_segmentation_direct_v2}
\end{table*}

\begin{figure}[htpb]
    \centering
    \includegraphics[width=0.48\textwidth]{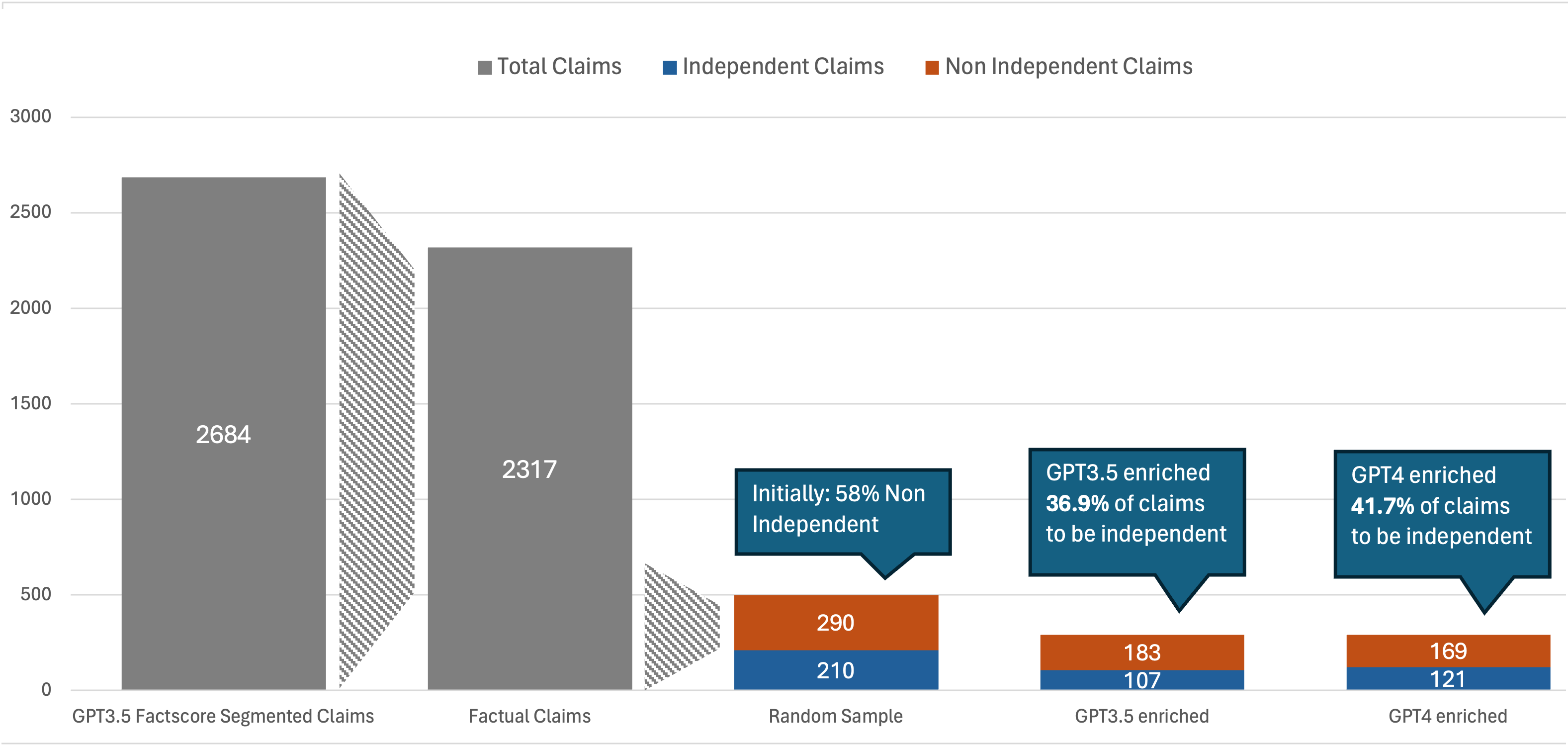}
    \caption{Statistics of contextualization of the 290 created claims by GPT3.5 and GPT4, evaluated by GPT4}
    \label{fig:claim_independence_enrichment}
\end{figure}

To test the enrichment, we utilize only the GPT-3.5 system, as described in Table \ref{tab:claim_segmentation_low_level}. From the 
2,317 unique and factual claims, as segmented by the original GPT-3.5 system, we take a random sample of 500 and assess their independence using the GPT-4 prompt from the previous step. We observe \textbf{290 out of 500} were deemed to be "not independent" by GPT-4. We then perform the enrichment by applying a one-shot prompt with both GPT-3.5 and GPT-4 to implement the function $f_{\text{enrich}}(q, a, c_{\text{non-independent}})$ and compare the results to the original claims. The comparison is conducted using the non-independence detection system described above. The quality of this step is measured in the reduction of non-independent claims. The results are presented in Figure \ref{fig:claim_independence_enrichment}.

The enrichment function managed to make an additional 107/290 with GPT-3.5 and 121/290 with GPT-4 claims independent, i.e., further reducing the number of non-independent claims by 36.9\% (GPT-3.5) and 41.7\% (GPT-4). This is a considerable improvement that increases the number of claims usable for later attribution steps. Nevertheless, many claims still remained without context. Another observation is that the enrichment has noticeably increased the average number of characters of the claims. Initially, the average number of characters for independent claims was 66.0 and 59.4 for non-independent claims. The revision by GPT-4 increased it to 155.6 characters, and the enrichment by GPT-3.5 to 145.9 characters. 
Later, we evaluate the impact of claim enrichment on the evidence retrieval process (Section \ref{sec:impact_retrieval}).

\subsubsection{Answer Segmentation with Context -- Direct Segmentation}\label{sec:claim_segmentation_direct}
An alternative to enrichment is directly segmenting the answer into multiple independent claims with context. This approach implements the function $f_{\text{segment}}(r, q)$ by using a one-shot prompt and GPT3.5 and GPT4 as LLMs. To evaluate the result quantitatively, we compare the average number of claims and the length of claims with those from alternative approaches to answer segmentation. This step is done on a subset of 100 question-answer pairs from ExpertQA. The prompt requests the model to print out a structured list of claims. The exact prompt can be found in Appendix \ref{sec:appendix-prompts}, Table \ref{tab:direct_seg}. 
The results are presented in Table \ref{tab:claim_segmentation_direct_v2}.

Upon applying the segmentation to the responses from GPT-4, an increase in the number of claims was observed, aligning with the levels obtained through the original FactScore segmentation. This implementation aims to diminish non-independence, given that the original FactScore segmentation relied on SpaCy sentences, which exhibited non-independence in 80\% of instances. As generating independent claims from non-independent inputs is not possible, employing GPT-4 as a baseline may mitigate this issue.

\begin{table}[htpb]
\footnotesize
    \centering
    \caption{Comparison of claim enrichment on the retrieval performance.}
    \begin{tabular}{lcccc}
    \hline
    \textbf{Model} & \textbf{Contr.} & \textbf{Entail.} & \textbf{No Rel.} \\ \hline
    Original \textbf{Independent} & 5.6\% & 42.2\%  & \textbf{52.2\%} \\
    Original \textbf{Non-Ind.} & 3.6\% & 24.1\% & \textbf{71.3\%} \\ 
    Enriched \textbf{Independent}  & 6.1\% & 35.4\%  & \textbf{58.6\%} \\
    Enriched \textbf{Non-Ind.}  & 1.3\% & 20.5\%  & \textbf{78.2\%} \\ \hline
    \end{tabular}
    \label{tab:claim_enrichment_retrieval}
\end{table}

\subsection{Factuality \& Independence}
\begin{figure}[htpb]
\centering
\includegraphics[width=0.49\textwidth]{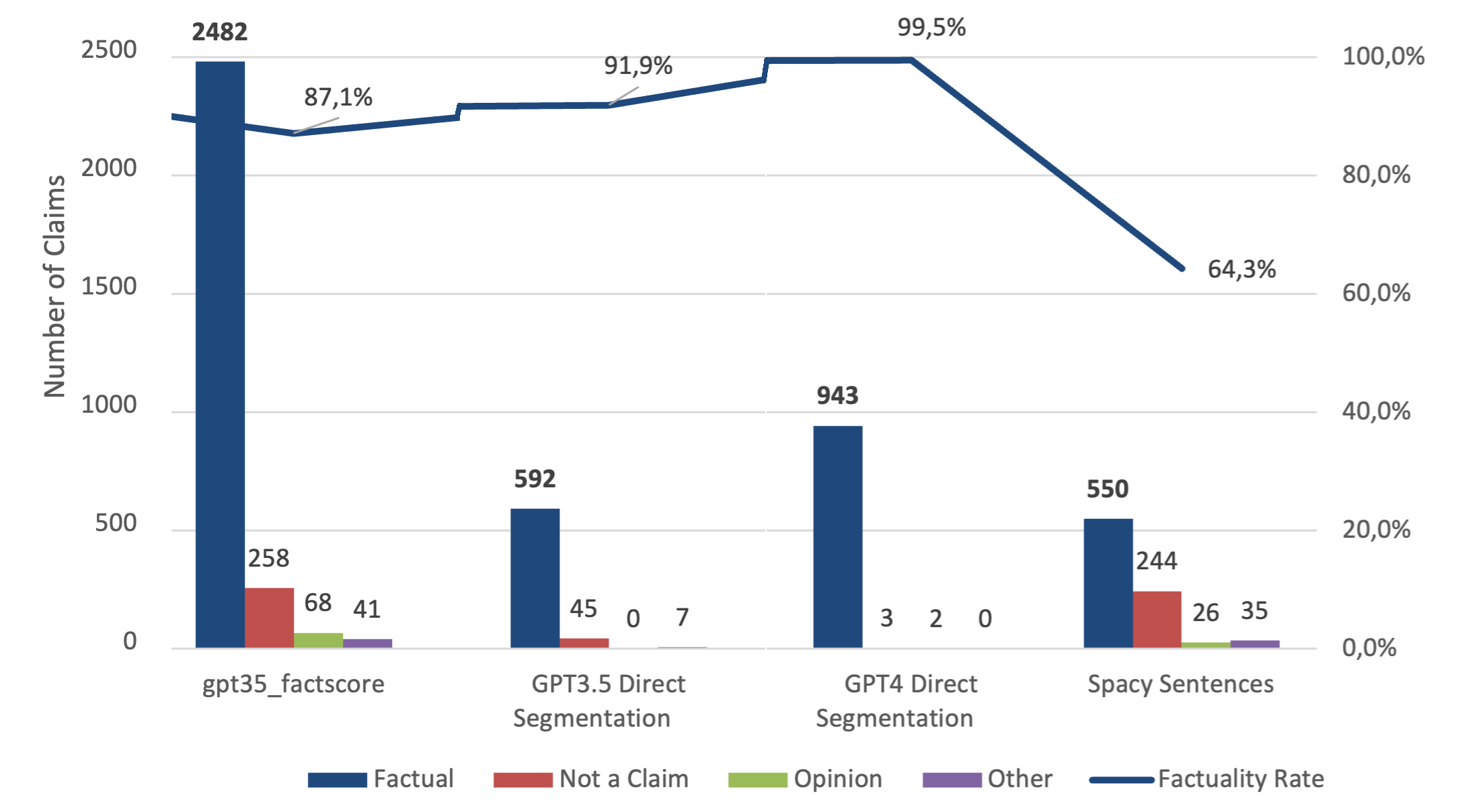}
\caption{Visualization of the factuality evaluation statistics for the four different systems}
\label{fig:claim_factuality_new}
\end{figure}

The next step in the evaluation involves analyzing the factuality of individual claims. This is done employing the same methodology as described in Section \ref{sec:claim_relevance_determination}, with previous results in Table \ref{tab:claim_segmentation_low_level}. The outcomes of the direct answer segmentation are depicted in Figure \ref{fig:claim_factuality_new}.

This figure clearly demonstrates an improvement in the factuality rate of the claims generated by both GPT-4 and GPT-3.5 compared to SpaCy sentence segmentation, with the factuality rate increasing from 64.3\% to 99.5\% for GPT-4 and to 91.9\% for GPT-3.5. These results suggest that this approach is a significantly better alternative to spaCy tokenization.

\begin{table*}[htpb]
\small
    \centering
    \caption{Comparison of direct answer segmentation on the retrieval performance (more Entailment is better).}
    \begin{tabular}{lcccc}
    \hline
    \textbf{Model} & \textbf{Contradiction} & \textbf{Entailment} & \textbf{Missing} & \textbf{No Relation} \\ \hline
    Original \textbf{Independent} & 5.6\% & 42.2\% & 0.0\% & \textbf{52.2\%} \\
    Original \textbf{Non-Ind.} & 3.6\% & 24.1\% & 2.4\% & \textbf{69.9\%} \\
    GPT3.5 Direct -- \textbf{Independent} & 4.2\% & 47.2\% & 0\% & \textbf{48.6\%} \\
    GPT3.5 Direct -- \textbf{Non-Ind.} & 0\% & 27.0\% & 2.7\% & \textbf{70.3\%} \\
    GPT4 Direct -- \textbf{Independent} & 0\% & 51.5\% & 0\% & \textbf{48.5\%} \\
    GPT4 Direct -- \textbf{Non-Ind.} & 2.0\% & 14.3\% & 2.0\% & \textbf{81.6\%} \\ \hline
    \end{tabular}
    
    \label{tab:claim_segmentation_direct_retrieval}
\end{table*}

\subsection{Impact on Evidence Retrieval}\label{sec:impact_retrieval}
The evaluation of the impact of claim enrichment on evidence retrieval is conducted using the same 2,317 (question, response, claim) triplets, which were classified by the GPT-3.5 system as factual, as in the previous setup. The retrieval process is conducted using the same GPT-3.5 enriched claim-based retrieval system 
For assessing the impact of claim enrichment on retrieval (function $f_{\text{enrich}}(q, a, c_{\text{non-independent}})$), we compare a sampled yet stratified set of claims across four categories: originally independent, originally non-independent, enriched (by GPT4)  non-independent, and enriched (by GPT4) independent claims.
The enriched claims are based on the originally non-independent claims. We utilize DeBERTa to evaluate the claim-evidence relation.

The findings are presented in Table \ref{tab:claim_enrichment_retrieval}. The table reveals several interesting findings: Firstly, it is evident that originally independent claims highly outperform originally non-independent claims in the evidence retrieval pipeline. 
Upon enriching the originally non-independent claims with GPT-4, as described in the previous section ($f_{\text{enrich}}(q, a, c_{\text{non-independent}})$), the claims that were successfully enriched show a big improvement in performance within the retrieval pipeline. This indicates that enriching (contextualizing) claims enhances the retrieval performance.
The successfully enriched claims approach the performance of the originally independent claims, with a "No Relation" share of 58.6\%. However, claims that were not successfully enriched exhibit worse performance than the originally non-independent claims, with a "No Relation" share of 78.2\%. 
Overall, the effect of claim enrichment is a 16.2 percentage point reduction (69.9 to 53.7) of claim-source pairs with no relation.

\begin{table*}[htpb]
\small
    \centering
    \caption{Comparison of different embedding models and context window splitters on the retrieval performance (more Entailment indicates better performance).}
    \begin{tabular}{lccc}
    \hline
    \textbf{Model} & \textbf{Contradiction} & \textbf{Entailment} & \textbf{No Relation} \\ \hline
    Ada 2.0  & 2.9\% & 41.0\%  & \textbf{56.0\%} \\
    AnglE  & 2.9\% & 39.5\% & \textbf{57.5\%} \\
    SBert + Recursive CW & 0.0\% & 22.1\% & \textbf{76.1\%} \\
    SBert Baseline (Macro) & 0.9\% & 35.7\% & \textbf{62.5\%}\\ \hline
    \end{tabular}
    
    \label{tab:claim_segmentation_direct_retrieval_comp}
\end{table*}

Additionally, we evaluate the impact of direct answer segmentation on the retrieval process. For that, we use the random sample of 40 (question, response, claim) triplets per direct segmentation system, as described in Section \ref{sec:claim_segmentation_direct}.
The results are presented in Table \ref{tab:claim_segmentation_direct_retrieval}. As above, we analyze the share of (claim, evidence) pairs that are classified as "Missing" or "No Relation" by DeBERTa; a lower share means a better retrieval process. The table shows the claims were yet again enhanced when compared to the previous enrichment approach. Direct segmentation by GPT-4 records a combined "Missing + No Relation" share of 48.5\% for independent claims and 81.6\% for non-independent claims. This represents a significant improvement for independent claims compared to both enriched and original claims.

To summarize the findings, it can be concluded that direct segmentation with context by GPT-4 significantly surpasses both the original and enriched claims and outperforms comparative methods in aspects of retrieval, time efficiency, and independent claim generation. It nearly matches the performance of GPT-4 in enriching non-independent claims regarding the creation of independent claims and surpasses it in the retrieval process at the macro level.

\subsection{Analysis of Evidence Retrieval}
As a final step, we briefly evaluate the evidence retrieval process itself, analyzing different embedding models and context window sizes. We utilize claims generated by GPT-4 Direct, as this system was shown to be the best performer in the previous steps. We use the same random sample of 40 questions. We modify two dimensions of the retrieval process: the embedding model and the context window splitter. Instead of Sentence-BERT, we employ OpenAI Ada 2.0, which provides embeddings from GPT-3.5, and AnglE-Embeddings \cite{li2023angleoptimized} from a pre-trained sentence-transformer model optimized for retrieval. Rather than using a simple sliding window approach, we implement a recursive text splitter with overlap to capture more relevant information.

The search engine (Google Search Custom Search Engine) remains unchanged. The results are presented in Table \ref{tab:claim_segmentation_direct_retrieval_comp}. The results demonstrate that the Ada 2.0 Embeddings with the fixed 512c-size context window splitter outperform the overall SBert baseline, which was used in our previous experiments and depicted the best performance. The AnglE embeddings, optimized for retrieval, also outperform the Sentence-BERT baseline but fall behind the GPT-based Ada 2.0 embeddings. Interestingly, the recursive context window splitter with SBert embeddings performs significantly worse than the fixed context window splitter. 

\section{\uppercase{Discussion}}
The evaluation of various attribution methods revealed that the main challenge lies in the precise retrieval of relevant evidence snippets, especially considering the complexity of the query or the intended user need. A crucial aspect of effective retrieval is in formulating claims for subsequent search in a way that they are atomic, independent, and properly contextualized. Additionally, addressing the shortcomings in answer segmentation and independence was essential for improving the attribution process. Segmenting answers into independent (contextualized) claims was most effectively done using GPT-4, yet it did not achieve an 80\% success rate. This indicates that a general-purpose language model might not be the best choice for this task and could be improved in the future by a more specialized and smaller model tailored specifically for this purpose. Future work could involve fine-tuning models for detecting non-independent claims and exploring alternative approaches for source document retrieval. Additionally, future research should focus on expanding the scope of embedding models and their context windows for semantic search of evidence.

\section{\uppercase{Conclusion}}
In this paper, we analyzed automated answer attribution, the task of tracing claims from generated LLM responses to relevant evidence sources. By splitting the task into constituent components of answer segmentation, claim relevance detection, and evidence retrieval, we performed a case analysis of current systems, determined their weaknesses, and proposed essential improvements to the pipeline. Our improvements led to an increase in performance in all three aspects of the answer attribution process. We hope our study will help future developments of this emerging NLP task.

\bibliographystyle{apalike}
{\small
\bibliography{example}}

\section*{\uppercase{Appendix}}

This is the appendix with additional material.

\subsection*{Technical Setup and Manual Annotation}
All GPT 3.5 and GPT 4 models were accessed through the official OpenAI API. Version \textit{turbo-0125} for GPT 3.5 and \textit{0125-preview} for GPT 4, or as indicated in the text. For local experiments (such as model embeddings with sentence transformers, DeBERTa entailment prediction, etc.), one A100 GPU with 40GB of VRAM was used, for a duration of one computation hour per experiment. No fine-tuning was performed by us, models like SegmenT5 and DeBERTa-v3 were used out-of-the-box, found in cited sources and HuggingFace.

Whenever we refer to manual annotation of data examples, this was done by two paper authors, who have a master's degree in computer science and are pursuing a PhD degree in computer science. None of the annotations required in-depth domain knowledge and were mostly reading comprehension tasks.

\subsection*{Prompts}
\label{sec:appendix-prompts}
The used prompts are given in Tables 10--14.

\begin{table*}[htpb]
\small 

\centering
\caption{Overview of applied prompts for GPT answer segmentation and claim relevance (check-worthiness) detection.}

\begin{tabular}{p{28mm}p{117mm}}
\hline
\textbf{Use Case} & \textbf{Prompt Content}\\
\hline

FactScore Answer Segmentation with GPT 3.5 & \scriptsize Please breakdown the following sentence into independent facts.

Don't provide meta-information about sentence or you as a system. Just list the facts and strictly stick to the following format: 

1. "Fact 1"

2. "Fact 2"

3. "..."

The sentence is:\\
 \hline

Claim Relevance / Check-Worthiness Detection & \scriptsize You are a factchecker assistant with task to identify a sentence, whether it is 1. a factual claim; 2. an opinion; 3. not a claim (like a question or a imperative sentence); 4. other categories.

Let's define a function named checkworthy(input: str).

The return value should be a python int without any other words, representing index label, where index selects from [1, 2, 3, 4].

For example, if a user call checkworthy("I think Apple is a good company.") 
You should return 2

If a user call checkworthy("Friends is a great TV series.") 
You should return 1

If a user call checkworthy("Are you sure Preslav is a professor in MBZUAI?") 
You should return 3 

If a user call checkworthy("As a language model, I can't provide these info.") You should return 4

Note that your response will be passed to the python interpreter, no extra words.

checkworthy("{input}")
\\ \hline

\end{tabular}
\label{tab:checkworth}
\end{table*}

\begin{table*}[h]
\centering
\caption{Overview of applied prompt for claim-evidence relation detection, i.e., entailment recognition (NLI) between the claim and retrieved evidence chunk with GPT 3.5.}

\begin{tabular}{p{25mm}p{120mm}}
\hline
\textbf{Use Case} & \textbf{Prompt Content}\\
\hline

Claim-Evidence Entailment Recognition & \scriptsize Your task is to determine if a claim is supported by a document given a specific question. Implement the function nli(question: str, claim: str, document: str) -$>$ str which accepts a question, a claim, and a document as input.

The function returns a string indicating the relationship between the claim and the document in the context of the question. The possible return values are:

"entailed" if the claim is supported by the document, 
"contradicted" if the claim is refuted by the document, 
"no\_relation" if the claim has no relevant connection to the document given the question. 

Your evaluation should specifically consider the context provided by the question. The output should be a single string value without additional comments or context, as it will be used within a Python interpreter.

Examples:

Question: "You are patrolling the local city center when you are informed by the public about a young girl behaving erratically near traffic. What are your initial thoughts and actions?"

Claim: "Trained professionals should handle situations like this."

Document: "Every trained professional football player should be adept at managing high-stress situations on the field."

Output: "no\_relation"

Question: "You are patrolling the local city center when you are informed by the public about a young girl behaving erratically near traffic. What are your initial thoughts and actions?"

Claim: "Trained professionals should handle situations like this."

Document: "Standard police officer training includes procedures for managing public disturbances and emergencies."

Output: "entailed" \\ 
\hline

\end{tabular}

\label{tab:gpt_nli}
\end{table*}

\begin{table*}[h]
\centering
\caption{Overview of applied prompt for the claim independence detection.}

\small

\begin{tabular}{p{25mm}p{120mm}}
\hline
\textbf{Use Case} & \textbf{Prompt Content}\\
\hline

Claim Independence Detection & \tiny You are tasked with determining whether a given claim or statement can be verified independently.
A claim is considered "independent" if it contains sufficient information within itself to assess its truthfulness without needing additional context or external information.
Your response must strictly be either "independent" or "not independent." Adhere to this format precisely, as your output will be processed by a Python interpreter.

Guidelines:

Evaluate if the claim provides enough detail on its own to be verified.
Do not consider external knowledge or context not present in the claim.
Respond only with "independent" if the claim is self-sufficient for verification; otherwise, respond with "not independent."
The below examples contain Rationales for explanation, which are not allowed in your response.

Examples:

Input: "The sun rises in the east."

Output: independent

Input: "Chemotherapy is no longer the recommended course of action."

Output: not independent

Rationale: The claim would require additional context, for example the type of cancer or the patient's medical history.

Input: "Opening up the aperture can overexpose the image slightly."

Output: independent

Input: "A young girl is running in front of cars."

Output: not independent

Rationale: The claim is situational and lacks specific details that would allow for independent verification.

Input: "They ensure the well-being of everyone involved."

Output: not independent

Rationale: The claim is vagues, it is not known who "They" are. \\
\hline

\end{tabular}
\label{tab:ind_prompt}
\end{table*}

\begin{table*}[h]
\centering
\caption{Overview of applied prompt for the claim enrichment process.}

\small

\begin{tabular}{p{25mm}p{120mm}}
\hline
\textbf{Use Case} & \textbf{Prompt Content}\\
\hline

 Claim Enrichment Prompt & \scriptsize Your task involves providing context to segmented claims that were originally part of a larger answer, making each claim verifiable independently.
 
This involves adding necessary details to each claim so that it stands on its own without requiring additional information from the original answer.
The claim should stay atomic and only contain one specific statement or piece of information. Do not add new information or more context than necessary!
Ensure that all pronouns or references to specific situations or entities (e.g., "He," "they," "the situation") are clearly defined within the claim itself.
Your output should consist solely of the context-enhanced claim, without any additional explanations, as it will be processed by a Python interpreter.

Example:

Question: "How to track the interface between the two fluids?"

Answer: "To track the interface between two fluids, you can use various techniques depending on the specific situation and the properties of the fluids. Here are a few common methods:

    ...
    
    4. Ultrasonic Techniques: Ultrasonic waves can be used to track the interface between fluids. By transmitting ultrasonic waves through one fluid and measuring the reflected waves, you can determine the position of the interface.
    
    ...
    
    It's important to note that the choice of method depends on the specific application and the properties of the fluids involved."
    
Claim: "Reflected waves can be measured."

Revised Claim: "Reflected waves can be measured to determine the position of the interface between two fluids." \\
\hline

\end{tabular}
\label{tab:enrich_prompt}
\end{table*}

\begin{table*}[h]
\centering
\caption{Overview of the prompt for direct claim segmentation with added context.}

\small 

\begin{tabular}{p{28mm}p{117mm}}
\hline
\textbf{Use Case} & \textbf{Prompt Content}\\
\hline

Direct Claim Segmentation with Context & \scriptsize Objective: Transform the answer to a question into its discrete, fundamental claims. Each claim must adhere to the following criteria:

Conciseness: Formulate each claim as a brief, standalone sentence.

Atomicity: Ensure that each claim represents a single fact or statement, requiring no further subdivision for evaluation of its truthfulness. Note that most listing and "or-combined" claims are not atomic and must be split up.

Independence: Craft each claim to be verifiable on its own, devoid of reliance on additional context or preceding information. For instance, "The song was released in 2019" is insufficiently specific because the identity of "the song" remains ambiguous. Make sure that there is no situational dependency in the claims.

Consistency in Terminology: Utilize language and terms that reflect the original question or answer closely, maintaining the context and specificity.

Non-reliance: Design each claim to be independent from other claims, eliminating sequential or logical dependencies between them.

Exhaustiveness: Ensure that the claims cover all the relevant information in the answer, leaving no important details unaddressed.

Strictly stick to the below output format, which numbers any claims and separates them by a new line. This is important, as the output will be passed to a python interpreter.

Don't add any explanation or commentary to the output.

Example:

Input:

Question: As an officer with the NYPD, I am being attacked by hooligans. What charges can be pressed?

Answer: If you're an NYPD officer and you're being assaulted by hooligans, you have the right to press charges for assault on a police officer, which is recognized as a criminal offense under New York law. Specifically, the act of assaulting a police officer is addressed under New York Penal Law § 120.08, designating it as a felony. Offenders may face severe penalties, including time in prison and monetary fines.

Output:

1. An NYPD officer assaulted by hooligans has the right to press charges for assault on a police officer. 

2. Assault on a police officer is deemed a criminal offense in New York. 

3. The act of assaulting a police officer is specified under New York Penal Law § 120.08. 

4. Under New York law, assaulting a police officer is categorized as a felony.

5. Conviction for assaulting a police officer in New York may result in imprisonment. 

6. Conviction for assaulting a police officer in New York may lead to monetary fines. \\
\hline

\end{tabular}
\label{tab:direct_seg}
\end{table*}

\end{document}